\documentclass[final,5p,times,twocolumn]{elsarticle}
\usepackage{amssymb}
\usepackage{amsmath}
\usepackage{booktabs}
\usepackage{multirow}
\usepackage{longtable}

\journal{arXiv}

\begin{document}

\begin{frontmatter}
\title{MicroNAS: An Automated Framework for Developing a Fall Detection System}
\author[first]{Seyed Mojtaba Mohasel}
\author[second]{John Sheppard}
\author[third]{Lindsey K. Molina}
\author[third]{Richard R. Neptune}
\author[fourth]{Shane R. Wurdeman}
\author[first]{Corey A. Pew}
\affiliation[first]{organization={Montana State University, Mechanical and Industrial Engineering},
            addressline={}, 
            city={Bozeman},
            postcode={}, 
            state={MT},
            country={USA}}
\affiliation[second]{organization={Montana State University, Gianforte School of Computing},
            addressline={}, 
            city={Bozeman},
            postcode={}, 
            state={MT},
            country={USA}}
            
\affiliation[third]{organization={Walker Department of Mechanical Engineering, The University of Texas at Austin},
            addressline={}, 
            city={ Austin},
            postcode={}, 
            state={TX},
            country={USA}}

\affiliation[fourth]{organization={Department of Clinical and Scientific Affairs, Hanger Clinic},
            addressline={}, 
            city={ Austin},
            postcode={}, 
            state={TX},
            country={USA}}

\begin{abstract}
This work presents MicroNAS, an automated neural architecture search tool specifically designed to create models optimized for microcontrollers with small memory resources. The ESP32 microcontroller, with 320 KB of memory, is used as the target platform. The artificial intelligence contribution lies in a novel method for optimizing convolutional neural network and gated recurrent unit architectures by considering the memory size of the target microcontroller as a guide. A comparison is made between memory-driven model optimization and traditional two-stage methods, which use pruning, to show the effectiveness of the proposed framework. To demonstrate the engineering application of MicroNAS, a fall detection system (FDS) for lower-limb amputees is developed as a pilot study. A critical challenge in fall detection studies, class imbalance in the dataset, is addressed. The results show that MicroNAS models achieved higher F1-scores than alternative approaches, such as ensemble methods and H2O Automated Machine Learning, presenting a significant step forward in real-time FDS development. Biomechanists using body-worn sensors for activity detection can adopt the open-source code  to design machine learning models tailored for microcontroller platforms with limited memory.
\end{abstract}



\end{frontmatter}




\section{Introduction}
\label{introduction}

Falls present a major health risk for individuals with lower limb amputation [1, 2]. Specifically, more than half of lower limb amputees report falling in the previous 12 months. Furthermore, of those reporting a fall, approximately 75\% report multiple falls [1]. Falls have the potential for multiple negative sequelae, including fractures, traumatic brain injuries, lacerations, sprains, hematomas, and even death [3]. More commonly, a fall may only result in minor injuries or bruises but can impact the person's confidence in their balance and mobility [3]. Consequently, they may limit their physical activity and social participation, leading to a decline in overall physical and emotional health.

 Falls also pose a barrier to successful rehabilitation, whether it be physical or emotional injury. The extent to which falls delay or prevent successful rehabilitation of individuals with lower limb amputations is unknown. Currently, evaluations of falling frequency are primarily conducted via survey and self-report. However, the accuracy of patient recall over an extended period is questionable [4, 5]. A more quantitative method is critically needed to evaluate the number of falls experienced in the amputee community to objectively evaluate the magnitude of the problem and the effects of specific interventions on fall risk. Multiple studies have developed simple, low-cost sensors that can be attached to an individual in order to detect fall events. The primary methodology for detecting falls is with the use of inertial measurement units (IMUs) that include some form of multi-axis accelerometers, gyroscopes and/or magnetometers. Wearable devices are uniquely suited for the amputee community as IMUs are often already incorporated into systems such as microprocessor knees and ankles [6]. Data from sensors are monitored to look for motion events that are classified as falls in comparison to regular activity. Simple fall detection algorithms utilize metrics such as threshold detection and orientation change to distinguish falls from everyday activity [7-9], however these often produce lower accuracy than required for real-world implementation. More complex algorithms are needed to accurately detect fall events with low false positive rates to provide clinically meaningful feedback for rehabilitation of individuals with lower-limb amputation.

 Machine learning (ML) techniques are often used to identify falls from raw sensor data [10]. ML has become a common technique in the field of biomechanics to identify activities of daily living (ADL) using body worn sensors [11]. In addition, integrated tools in MATLAB, Python, and other software have made it easy to create ML models with any data set. However, not all biomechanists are trained in appropriate ML techniques and can easily neglect best practices when developing ML models [12]. 

 Here we identify three main principles often overlooked when developing fall detection classifiers: 

 \textbf{\textit{Data Division:}} leave-one-participant out is often the preferred approach in medical settings. To ensure real-world applicability, data from different participants should be individually divided into training, validation, and test sets [12]. The training set determines model parameters (weights for neural networks and splitting rules for decision trees). The validation set is used to monitor the model's performance on data it has not seen during training. The validation set is used to identify model hyperparameters (filter size, number of filters, and learning rate for a neural network and maximum depth for a decision tree). The test set assesses the model's performance and estimates its real-world capabilities and should only include data from an individual participant that is not a part of the training and validation sets (entirely unseen). Comparisons between different ML methods must utilize consistent sets (train, validation, test) and apply appropriate statistical comparisons [13].

 \textbf{\textit{Class Imbalance:}} Utilizing ML models that assume equal sample distribution across classes is inappropriate as the number of ADL samples significantly outnumbers fall samples leading to poor classification performance for the minority class. This issue is commonly referred to as class imbalance, and it is quantified using the imbalance ratio (IR), the number of majority samples divided by the number of minority samples [14]. Models specifically designed to handle class imbalance or methods capable of addressing class imbalance must be used [15]. 

 \textbf{\textit{Application:}} In resource-constrained environments (e.g., when using microcontrollers), researchers encounter hardware limitations for run time memory capacity [10]. Attaining high sensitivity and specificity for fall detection without addressing the capabilities of microcontrollers that will run the ML model is insufficient. A critical consideration is the model run-time size, as microcontrollers have limited memory. Researchers should develop models that account for this constraint or utilize techniques like compression [16], quantization [17], and pruning [18] to reduce the model size and then report the appropriate performance metrics.

 Inattention to these principles may result in misuse of ML techniques. Models may demonstrate excellent performance on tested data from a laboratory setting but may not be feasible for deployment and generalize poorly to real-world environments. While biomechanists can work with ML experts to navigate best practices, the interactive process can be time-consuming and require significant effort from both parties [19]. In addition, since each dataset has unique characteristics, the development of diverse methods, including neural networks and Ensemble models [20], is essential to find the best fall detection model.

 Automated machine learning (AutoML) can address these challenges by automating the entire ML pipeline [21]. AutoML is a framework designed to automate the application of ML to real-world problems [19]. AutoML frameworks streamline various stages of the ML pipeline development, including data preprocessing, model selection, hyperparameter tuning, and performance evaluation. Introducing an AutoML framework, particularly in the context of biomechanical systems like fall detection, would represent a significant step forward in making advanced technology accessible to non-experts. With automated tools, biomechanists can focus more on the experimental design, data collection, results analysis, and clinical applications of fall detection, while spending less time on the complexities of model development.

 AutoML leverages currently available, advanced algorithms to empower biomechanists to build ML applications appropriately, without requiring extensive statistical or ML knowledge [22]. However, AutoML tools have limitations: (a) H2O AutoML, Auto-Sklearn, and AutoGluon do not handle multivariate IMU time-series data; (b) Akkio and Azure may support IMU data, but they require a subscription; and (c) Mcfly [23] handles time-series data but does not offer models optimized for resource-constrained environments. Therefore, current AutoML tools are not well-suited to address the unique requirements of developing an FDS.

 Deep Learning (DL) models are the leading approaches for an FDS [24]; however, automating the entire pipeline of a DL model for a microcontroller is challenging. For example, automating a DL pipeline requires expertise in signal processing, neural networks, and optimization [25]. In addition, DL models involve complex algebraic operations, leading to high power consumption and long response times when executed on wearable devices with limited resources [26]. These challenges have sparked the evolution of modern DL model training called micro controller unit-aware neural architecture search (MCU-aware NAS). MCU-aware NAS, a subfield of AutoML, has been developed to create model architectures compatible with small memory footprints of microcontrollers [27]. Despite the initial advances of MCU-aware NAS, it is still in its infancy, and model architecture design specifically tailored for microcontrollers is scarce [27]. The present work seeks to address several knowledge gaps in this domain.

 \textbf{Knowledge Gap 1: An open-source AutoML pipeline that handles time series data, class imbalance, and memory constraints in a microcontroller-based FDS is needed. }

 One AutoML approach for creating small memory footprint models is Tiny-NAS [27]. The principle behind TinyNAS is that higher Floating-Point OPeration per Second (FLOPS) architectures produce models with higher accuracy. FLOPS represent the number of additions, subtractions, multiplications, and divisions that a computer can perform in one second. TinyNAS employs two distinct stages to identify the best model within a given space. In stage 1, the design space [28, 29] is analyzed by sampling architectures, resulting in the selection of a part of the design space that accommodates higher FLOPS while meeting the memory requirements [27]. In stage 2, TinyNAS selects the best model by optimizing the memory size through pruning [27]. Pruning results in a smaller model size facilitating faster inference speed, lower memory requirements, and improved efficiency on resource-constrained devices. However, there is a tradeoff between model pruning and F1-score [30]. Improved NAS designs are essential to enable the deployment of KB-sized neural networks without sacrificing F-score on resource-limited microcontrollers [31]. Searching architectures by expanding the model first and pruning it afterward in two stages can reduce the F1-score performance depending on the pruning rate [30, 32]. Therefore, considering one comprehensive stage for model creation from the beginning is potentially a more efficient search strategy. 

 \textbf{Knowledge Gap 2: A more direct and comprehensive, single-stage approach is currently not available to optimize architectures for specific microcontroller deployment without pruning.}

 The primary objective of this work was to develop an automated ML framework called MicroNAS. Our secondary objective was to develop a fall detection system for individuals with lower limb amputation. We developed MicroNAS and tested it by developing a fall detection algorithm as a test case. MicroNAS is a custom neural architecture search tool that efficiently explores feasible space and creates deployable models using the microcontroller's memory size as a guide. In this work, we focused on utilizing the Espressif Systems 32-bit SoC (ESP32) [33], a microcontroller known for its cost-effectiveness, low power consumption with compact board size, facilitating its incorporation into a lower limb prosthesis. Our emphasis was on ESP32-S2 series with 320 KB runtime memory. The body-worn FDS in this work is specifically designed for the amputee community and utilizes a single sensor placed on the pylon of a lower-limb prosthesis. The FDS leverages AutoML techniques for neural network model creation, alleviating the burden on researchers to develop models manually. Ultimately, the MicroNAS tool can identify a suitable architecture for neural network deployment on low-cost microcontrollers such as the ESP32 [33]. 

 To evaluate MicroNAS's ability to address the knowledge gaps, we introduce the following two hypotheses.

 \textit{Hypothesis 1:} Neural network models optimized with MicroNAS achieve higher performance, based on F1-score, compared to ensemble models or open-source AutoML tools capable of handling class imbalance automatically and operating with minimal memory requirements for ESP32.

 \textit{Hypothesis 2:} Neural network model architectures generated by MicroNAS, without pruning, will achieve higher F1-scores and lower memory sizes compared to neural network models generated by NAS using magnitude-based weight pruning [18] to fit the memory constraint for ESP32.

\section{Methods}
 An experimental human subject protocol was performed to provide laboratory-based examples of activities of daily living (ADLs) and falling and provide the basis for the ML models. Next, an AutoML pipeline was developed to process and optimize the model creation process.

\subsection{Experimental Data Collection}

 IMU sensors (XSens, Enschede, Netherlands, 100 Hz recording) were placed proximally on the shanks (Figure 1) of 30 control participants with no lower limb amputation and 5 individuals with lower limb amputation. Table A1 in the Appendix contains the details of demographic information for each participant. In a laboratory setting, participants performed ADLs (walking, running, turning, sitting/standing from a chair, lying down/rising from a bed, picking an item from the floor, and ascending/descending stairs and inclines) as well as simulated falls (forward, backward, left/right lateral, trip and recovery). All activities of daily living were grouped with the label 'ADL' and all fall types were grouped as `Fall' during model training for the FDS.

 \begin{figure}[ht]
     \centering
     \includegraphics[width=0.3\linewidth]{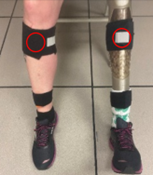}
     \caption{IMU sensors on both shanks of lower limb amputee, red circles indicate IMU locations. }
     \label{fig:1}
 \end{figure}

\subsection{MicroNAS Development}

 Details on data division, model selection, data segmentation, creation of MicroNAS and its functions, and ML data processing are presented below.

\subsubsection{Data Division}
 Data division created different Training, Validation, and Test data sets by shuffling data from each participant. The Test set consisted of a single participant. Each participant was in the Test set once creating 35 unique test data sets for evaluating model performance and statistical testing. The remaining participants were then divided into Training and Validation sets (90\% and 10\% of the remaining data, respectively) [34]. For each of the 35 combinations, data from the single Test participant were first removed, using leave-one-participant-out [35]. The Validation set always consisted of three amputee participants and three control participants, chosen using Uniform Randomization from each group (Python 3.10 random library). The remaining participant data was then used for Training (Figure 2). Training data included IMU signals from both shank sensors, boosting data volume. Data for Validation and Test only included from the sensor on the prosthetic limb for amputees and the non-dominant limb for control participants. 

 \begin{figure*}[ht]
     \centering
     \includegraphics[width=1\linewidth]{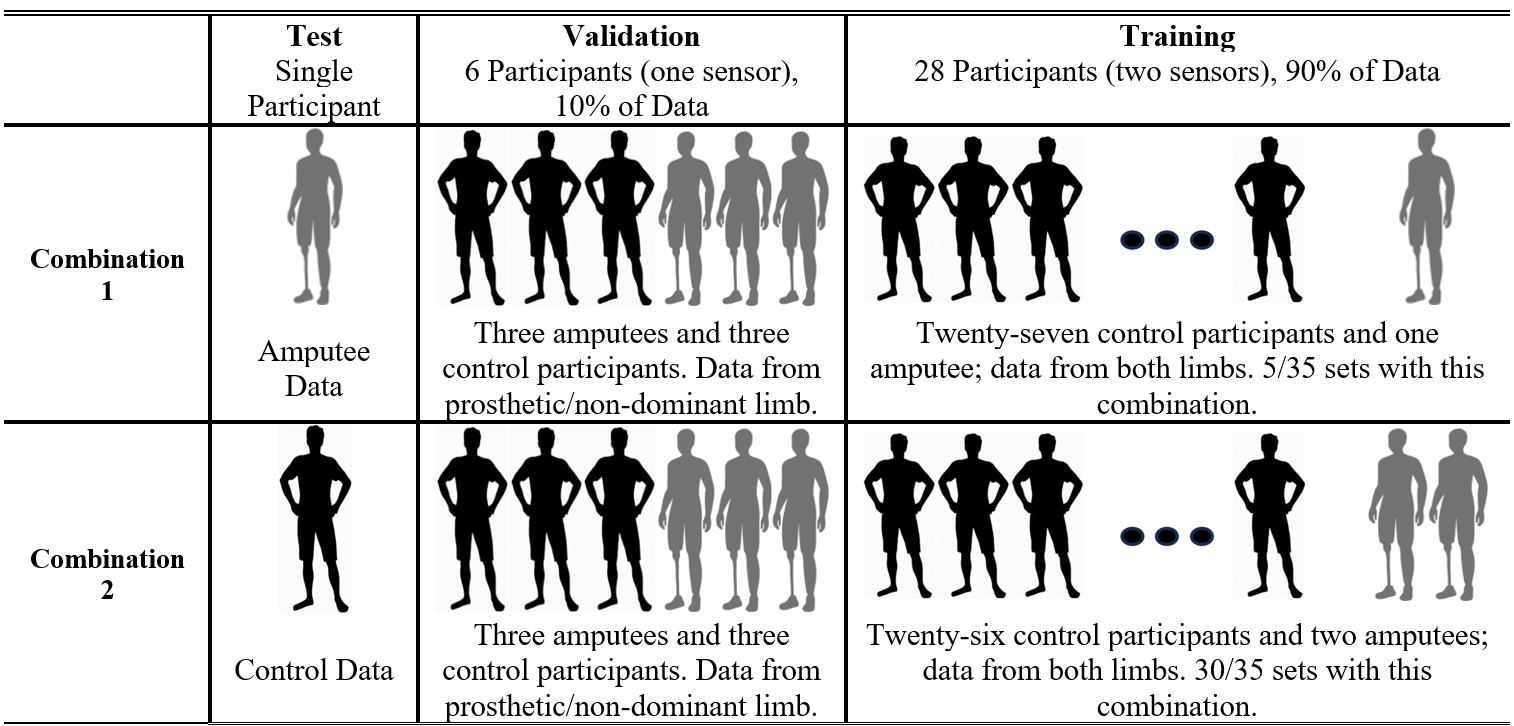}
     \caption{Data division into Training, Validation, and Test sets. The Test set consisted of a single participant with a lower limb amputation (Combination 1) or no lower limb amputation (Combination 2). Each participant was in the Test set once creating 35 unique test data sets for evaluating model performance and statistical testing. }
     \label{fig:2}
 \end{figure*}

\subsubsection{Class imbalance and model selection}
 Falling happens infrequently compared to daily activities, with this dataset containing 98.3\% ADL and only 1.7\% Fall classes. To handle class imbalance in neural networks, we employed a weighting method [36] that assigns higher weights to the fall class, with weights ($w_i$) defined as:
\begin{equation}
    w_i\mathrm{=}\frac{n}{kn_i}
    \label{EQ1}
\end{equation} 
where the total number of windows (section 2.2.3) (\textit{n}), the number of windows in each class ($n_i$), and the number of classes (\textit{k}) (in our case \textit{k} is 2) determine the weight ($w_i$). These weights are then considered in binary Cross-Entropy loss shown by equation \eqref{EQ2} during the training phase. Cross-entropy loss is a commonly used loss function in classification problems that calculates the difference between the predicted probability distribution and the actual probability distribution of a set of classes (Fall versus ADL). During the backward propagation phase of the training process, the gradient of the binary cross-entropy loss with respect to the model's parameters (weights and biases) is calculated and used to update the model's parameters. Weighted cross entropy corresponds to:
\begin{equation}
    {{CE=w}_1t}_1{\mathrm{log} \left(f{\left(s\right)}_1\right)+{w_2t}_2{\mathrm{log} \left(f{\left(s\right)}_2\right)\ }\ }
    \label{EQ2}
\end{equation}
where $w_1$and $w_2$ represent the weight assigned to each class ($i$) to account for class imbalance, $t_1$and $t_2$ represent the true label for class $1\ \mathrm{and}\ 2$, and $f{\left(s\right)}_1$ and $f{\left(s\right)}_2$ represent the predicted probability or score assigned by the model to class $1$ and $2$ for a particular input sample $s$.

 Neural network models studied included a 1D Convolutional Neural Network (1D CNN) [16] and a Gated Recurrent Unit (GRU) [37], both efficient in handling time series data. 1D CNN and GRU transform the representation of input data by feature extraction and make predictions based on features. 1D CNNs utilize filters to conduct convolution operations that capture temporal hierarchies in the segmented data (Section 2.2.3). GRUs are equipped with gates that regulate information flow, enabling them to capture temporal dependencies and retain past information in the segmented data.

 Ensemble models designed to address class imbalance, RUSboost [38] and EasyEnsemble [39], were selected. The ensemble models' performance were compared to neural networks (Hypothesis 1). Finally, H2O-AutoML was selected as the baseline for comparing our developed models. H2O-AutoML is an ensemble of various ML models and outperforms other AutoML tools on similar problems [40]. 

\subsubsection{Data segmentation}
 For neural network models, we segmented the raw data (3 axis accelerometer and gyroscope from the IMU) into sliding windows [41] to capture the dynamic nature of activities (Figure 3). Fall duration was approximately 1 second across all participants and pre-fall motions can often indicate an incoming fall [42], so we considered a window size of 1.2 seconds. We considered 90\% overlap for consecutive windows to identify the shifts between activities and provide large volume data for neural network's training phase.

 \begin{figure*}[ht]
     \centering
     \includegraphics[width=0.80\linewidth]{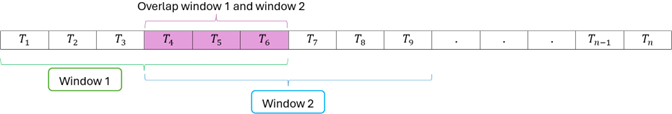}
     \caption{Windowing with overlap. T indicates the time stamp where T1 is the sample recorded at 0.01 seconds and T2 at 0.02 seconds during 100 Hz recording (Figure does not show time windows to scale). }
     \label{fig:3}
 \end{figure*}

\subsubsection{MicroNAS}
We designed MicroNAS, a neural architecture search tool, with a focus on identifying high performance (F1-score) architectures tailored for memory-constrained microcontrollers. MicroNAS development is organized into two key subsections: Algorithm and Search Space.

2.2.4.1 Algorithm:
Figure 4 outlines the flowchart of MicroNAS. MicroNAS utilizes the run time memory size and the search space as inputs, creating models that can be deployed on microcontrollers (e.g., the ESP32 hardware with 320 KB runtime memory sizes). Random search [43] maximizes the objective function (F1-score) by simultaneously exploring various layer sequences and hyperparameters (Section 2.2.4.2 Tables 1 and 2). Keras Tuner [44] in Python, a library developed for tuning hyperparameters of Neural network models, was used to implement the Random Search. Once Random Search determines an architecture with specific hyperparameters that meets the memory constraint, Adaptive Moment Estimation (Adam) [45] iteratively minimizes the cross-entropy loss function (Equation 2) of the determined architecture. Adam utilizes an extension of stochastic gradient descent with backward propagation [46] to update the architecture's parameters, including weights and biases, over a series of epochs. We adopted a fixed-budget approach [47], developing 20 architectures that meet the memory constraint as our termination criterion This strategy was implemented to both provide our models with opportunities for enhancing their F1-scores and ensure the completion of testing our two hypotheses within a month, on available hardware. The execution took place on Tempest, a high-performance computing research cluster at Montana State University, utilizing 25GB of RAM and a single Nvidia A100 GPU.

\begin{figure}[ht]
    \centering
    \includegraphics[width=0.75\linewidth]{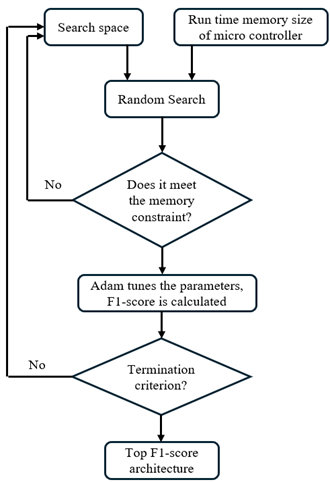}
    \caption{Model development and stages in MicroNAS }
    \label{fig:4}
\end{figure}

2.2.4.2 Search Space: Categorized into two areas: 1) architecture structure (layer sequences) and 2) hyperparameters. MicroNAS defines a three-dimensional search space (depth, width, and temporal resolution [48]) with diverse layer combinations, to provide the largest search space possible to find a memory-efficient architecture for the time series data.

\textbf{Architecture structure (layer sequences):} Table 1 presents the order of different layer types in our 1D CNN and GRU specific pipelines. Each pipeline begins with a batch normalization layer. In the 1D CNN pipeline, the first block comprises a convolutional layer and optional batch normalization and pooling layers. A second block, identical to the first, is optional and can be repeated based on the microcontroller's memory constraint (up to 10 times allowing the creation of networks with varying depths). Preceding the final block, there is an optional selection of global average pooling, dropout layer, and a fixed flatten layer. The last block (can be repeated up to 3 times) includes optional batch normalization layer, dropout layer, and fully connected layers with different number of neurons for dropout and fully connected (table 2) allowing the creation of networks with varying widths. The GRU pipeline resembles the 1D CNN pipeline, with the distinction that it incorporates a GRU layer instead of a convolutional layer. The described layer sequences facilitate the exploration of various architecture structures including combinations of Convolution, Pooling, Dropout, Batch Normalization similar to previous human activity recognition systems [49, 50, 51, 52, 53].

\begin{table}[ht]
  \caption{Architecture (flow of data through the network) of developed pipelines.}
\label{tab:1}
 \centering
 \resizebox{\linewidth}{!}{%
\begin{tabular}{@{}llll@{}}
\toprule
\textbf{Block Name}                                                     & \textbf{Layer type}    & \textbf{Option} & \textbf{Range}                                                                                                   \\ \midrule
                                                                        & Batch normalization    & Yes             & 1                                                                                                                \\
\multirow{3}{*}{\begin{tabular}[c]{@{}l@{}}CNN\\ or\\ GRU\end{tabular}} & Convolution or GRU     & Yes             & \multirow{3}{*}{\begin{tabular}[c]{@{}l@{}}Append or remove\\ random in the range\\ of {[}1; 10{]}\end{tabular}} \\
                                                                        & Batch normalization    & Yes/no          &                                                                                                                  \\
                                                                        & Pooling (CNN Only)     & Yes/no          &                                                                                                                  \\ \midrule
\multirow{3}{*}{Pre-final}                                              & Global average pooling & Yes/no          & \multirow{3}{*}{1}                                                                                               \\
                                                                        & Dropout layer          & Yes/no          &                                                                                                                  \\
                                                                        & Flatten layer          & Yes             &                                                                                                                  \\\midrule
\multirow{3}{*}{Final}                                                  & Batch Normalization    & Yes/no          & \multirow{3}{*}{\begin{tabular}[c]{@{}l@{}}Append or remove\\ random in the range\\ of {[}1; 3{]}\end{tabular}}  \\
                                                                        & Fully Connected        & Yes/no          &                                                                                                                  \\
                                                                        & Dropout                & Yes/no          &                                                                                                                  \\ \midrule
                                                                        & Fully Connected        & Yes             & 1                                                                                                                \\ \bottomrule
\end{tabular}
}
\end{table}
 \textbf{Hyperparameters:} Table 2 details the hyperparameters utilized by MicroNAS. Hyperparameter base ranges were determined by existing literature to encompass the widest range of usable space.

\onecolumn
\begin{longtable}{p{2.6cm}p{3.5cm}p{6cm}p{1.5cm}p{1.5cm}}
\caption{Selected hyperparameters for tuning in neural network models. The information within parentheses indicates that a specific hyperparameter is applicable to a particular}
\label{tab:2}\\
\toprule
\textbf{Hyperparameter}                         & \textbf{Definition}                                                                                                                                                          & \textbf{Purpose}                                                                                                                                                 & \textbf{Value}                                                                        & \textbf{References}                          \\* \midrule
\endfirsthead
\multicolumn{5}{c}%
{{\bfseries Table \thetable\ continued from previous page}} \\
\toprule
\textbf{Hyperparameter}                         & \textbf{Definition}                                                                                                                                                          & \textbf{Purpose}                                                                                                                                                 & \textbf{Value}                                                                        & \textbf{References}                          \\* \midrule
\endhead
\bottomrule
\endfoot
\endlastfoot
Number of Filters (CNN)                         & Number of filters in a convolutional layer                                                                                                                                   & Controls model depth and complexity. More filters increase complexity and extract additional features. Related to width of the search space.                     & 1$\mathrm{\sim }$500                                                                  & {[}69{]} {[}70{]} {[}71{]} {[}72{]} {[}73{]} \\\midrule
Filter Size (CNN)                               & Size of the filters in a convolutional layer                                                                                                                                 & Determines feature extraction's receptive field. Related to temporal resolution of search space.                                                                 & 1$\mathrm{\sim}$ 8                                                                    & {[}69{]} {[}70{]} {[}71{]} {[}72{]} {[}73{]} \\\midrule
Filter Stride (CNN)                             & Number of samples the filter omits during input convolution                                                                                                                  & Reduced memory utilization and increased inference computation                                                                                                   & 1,2                                                                                   & {[}70{]}{[}73{]}                             \\\midrule
Padding                                         & Number of zero value time steps added to each side of the input data. Applied to input data before convolution.                                                              & Preserves spatial dimensions and prevents information loss at the edges of windows to increase performance.                                                      & Yes, no                                                                               & {[}74{]}                                     \\\midrule
Activation Function                             & Applied to the output of a neuron. Rectified Linear Unit function, hyperbolic tangent, and sigmoid activation functions.                                                     & Enables non-linear learning                                                                                                                                      & ReLu, Tanh, Sigmoid                                                                   & {[}75{]}{[}76{]}{[}77{]}                     \\\midrule
Pooling (CNN)                                   & Down sampling operation to reduce spatial dimensions of an input data by returning the maximum or average of the arrays of input data.                                       & Reduces computation and extracts features at different scales. Improves memory efficiency.                                                                       & Pool length: 2, 4,8,16 Max or Average                          & {[}70{]}{[}75{]}{[}72{]}                     \\\midrule
GRU size (GRU)                                  & Dimensionality of the learned representation influences the model's ability to capture and learn patterns in sequential data                                                 & Determines the dimensionality of the output space                                                                                                                & 30$\mathrm{\sim}$256                                                                  & {[}69{]}{[}72{]}                             \\\midrule
Number of Neurons in Each Fully connected Layer & Connects neurons from the previous layer to the current layer.                                                                                                               & More neurons increase model complexity and memory requirement                                                                                                    & 3$\mathrm{\sim }$1024                                                                 & {[}76{]}{[}69{]}{[}70{]}\newline{[}71{]}{[}74{]}     \\\midrule
Batch size                                      & Number of windows that are processed together                                                                                                                                & Improves computational efficiency in the training phase.                                                                                                         & 916                                                                                   & {[}75{]}{[}76{]}\newline{[}70{]}{[}78{]}             \\\midrule
Epoch                                           & One epoch is when the model processes all the batches in the training phase.                                                                                                 & Ensures that the neural network learns from the entire training dataset in an iterative manner. For this dataset, 100 is effectively no upper limit.             & 100                                                                                   & {[}79{]}                                     \\\midrule
Early stopping                                  & If the Cross-Entry loss does not decrease between a set number of epochs, training was terminated. The models is saved at the point of the last Cross-Entropy loss decrease. &     Prevents overfitting and improves generalization to unseen data. & 10                                                                                    & {[}72{]}                                     \\\midrule
Dropout                                         & Temporarily ignoring neurons (excluding output neurons) during training for distribution of weighting.                                                                       &                                                                                                                                                                  & P = 0.2$\mathrm{\sim }$0.3 (GRU) P=0.4$\mathrm{\sim}$0.5 (CNN) & {[}70{]} {[}71{]}{[}74{]} {[}78{]}           \\\midrule
Initial Learning Rate                           & Starting magnitude for the optimization step of the stochastic gradient descent algorithm which influences convergence speed.                                                &                                                                                                                                                                  & 0.000001$\newline \mathrm{\sim }$0.01                                                          & {[}75{]}{[}69{]}{[}76{]}\newline{[}80{]}{[}81{]}     \\\midrule
Lasso Regularization                            & Adds a penalty term to the Cross-Entropy loss function. Encouraging sparsity in the weights of the dense layers.                                                             &                                                                                                                                                                  & 10$\mathrm{\wedge}$-5 $\mathrm{\sim }$ 10$\mathrm{\wedge}$-3                          & {[}82{]}                                     \\\midrule
Optimizer                                       & Adaptive Moment Estimation (Adam) is an extension of stochastic gradient descent. It is selected due to its good convergence and speed quality                               & Tunes the parameters (weights and biases) of the neural network                                                                                                  & Adam                                                                                  & {[}78{]}{[}81{]}{[}70{]}                     \\* \bottomrule
\end{longtable}

\twocolumn

 \subsubsection{Ensemble data processing: }RUSBoost and EasyEnsemble areensemble models that provide a baseline for comparison with the MicroNAS developed neural network models. For ML model development, random search tuned the hyperparameters listed in Table 3. The ensemble models were trained on raw data in the Training set. In the Validation and Test set, we utilized a majority vote approach, involving the smoothing [54] of Fall/ADL predictions across 120 predictions to improve the F1-score of predictions. Ensemble models make predictions for 120 samples (using majority vote on a window) with 90\% overlap for the next predictions (Figure 5) to make them comparable with the neural network models (Figure 3). 

\begin{figure*}[ht]
    \centering
    \includegraphics[width=0.80\linewidth]{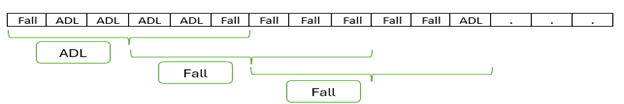}
    \caption{Windowing on Ensemble model predictions. Each prediction is for each time step and majority voting is used on 120 consecutive predictions. }
    \label{fig:5}
\end{figure*}

\begin{table*}[ht]
    \centering

 \caption{Selected hyperparameters for tuning in Ensemble models }
\label{tab:3}
\begin{tabular}{p{2.5cm}p{5cm}p{3cm}p{2.5cm}p{1.5cm}}
\toprule
\textbf{Hyperparameter} & \textbf{Definition}                                         & \textbf{Purpose}                          & \textbf{Values}            & \textbf{References} \\ \midrule
Number of estimators    & The number of trees in the forest.                          & Reduces overfitting and improve F1-score. & (10,250)                   & [69] [83] [84]      \\
Max depth               & The maximum depth of each decision tree.                    &                                           & (2,46)                     & [69][84]            \\
Learning rate           & The rate at which the model learns from the data.           &                                           & (0.001,1)                  & [69][83]            \\
Max features            & The maximum number of features each tree is allowed to use. &                                           & [sqrt, log2, None] &                     \\ \bottomrule
\end{tabular}
\end{table*}

 \subsection{Hypothesis testing:} We considered the F1-score of positive class (fall) as our optimization metric and for our hypothesis testing, as we were solely interested in the performance of the fall detector. F1-score (equation 3) is the harmonic mean of precision (which reduces false alarms) and recall (which increases fall detection). F1-score provides a single metric to evaluate the performance of neural networks and ensemble models focused on the minority class [55].  
\begin{equation}
    F1\ Score=\mathrm{2}\times \frac{Precision\ \times \ Recall}{Precision+Recall}
\end{equation}
Precision is the ratio of correctly predicted positive windows (Falls) to the total predicted positives. Precision determines the accuracy of the fall predictions, specifically, it measures the proportion of windows that were correctly identified as falls out of all windows predicted as falls.
\begin{equation}
    Precision=\frac{True\ Pos}{True\ Pos\mathrm{+}False\ Pos}
\end{equation}
Recall is the ratio of correctly predicted positive windows to all actual windows in the positive class. It measures the ability to identify falls out of all actual falls.
\begin{equation}
    Recall\ \left(Sensitivity\right)=\frac{True\ Pos}{True\ Pos\mathrm{+}False\ Neg}
\end{equation}
Low precision results in high false alarms and low recall indicates missing actual falls. All introduced metrics will be reported for our developed models. These metrics only provide information about the fall class. In addition, the macro-averaged precision, recall, and F1-score are reported to reflect the overall performance of our models. Macro-averaging treats each class (Fall/ADL) equally by calculating the metric for each class independently and then taking the average [56]. Macro-averaging considers ML performance in both Fall and ADL classes, irrespective of their frequencies.

 \textbf{\textit{Hypothesis 1:}} Neural network models optimized with MicroNAS achieve higher performance, based on F1-score, compared to ensemble models or open-source AutoML tools capable of handling class imbalance automatically and operating with minimal memory requirements for ESP32.  

 Hypothesis 1 compares the F1-scores of two neural network models (1D CNN and GRU), two ensemble models (RUSBoost and EasyEnsemble), and one open-source AutoML method (H2O AutoML) with each method generating an F1-score for each participant. Comparisons were made between models to determine differences. Parametric tests (ANOVA, t-test) have assumptions of Normality, equal variances, and independence. The Shapiro-Wilk test [57] was used to evaluate if the 35 F1-scores followed a normal distribution. The Levene's test [58] was used to assess the equality of variances. Since F1 scores are generated by different ML models, the independence assumption is validated. The Shapiro-Wilk test indicated that the F1-score samples for our models did not follow a normal distribution. Therefore, we used the Kruskal-Wallis test [59], a nonparametric version of ANOVA followed by the Wilcoxon Signed-Rank test [60], a non-parametric test, with Bonferroni correction to determine differences between the five models tested. 

 \textbf{\textit{Hypothesis 2:}} Neural network model architectures generated by MicroNAS, without pruning, will achieve higher F1-scores and lower memory sizes compared to neural network models generated by NAS using magnitude-based weight pruning [18], to fit the memory constraint for ESP32. 

 We selected 1D CNN as its hyperparameters (number of filters, filter size, stride size, and pooling size) provide more flexibility compared to GRU for pruning and our hypothesis testing. MicroNAS considers the memory of the microcontroller from the beginning. In contrast, NAS does not consider the constraint on memory and focuses on finding the highest F1-score model in stage 1. Then in stage 2, it prunes the model to fit the memory of ESP32. The Wilcoxon signed rank test compares the F1-scores and memory sizes of the 1D CNN models from MicroNAS and pruned 1D CNN created by NAS.

 \textbf{NAS pruning setup:} Initially we started with imposing no constraint on memory for NAS and iteratively reduced the memory size (up to 2400KB) until magnitude-based weight pruning could create a model that fit the memory of ESP32. Pruning reduces the memory size of our model by gradually reducing the number of parameters. It utilizes initial and final sparsity as hyperparameters in an optimization loop to incrementally remove less important weights during training.  Sparsity refers to the property where a subset of the model parameters (weights) have a value of exactly zero. Initial sparsity is the fraction of weights set to zero at the beginning of the pruning process. The weights with the smallest absolute values are usually the first to be pruned because they have the least effect on the network's output. Final sparsity is the targeted fraction of weights to be zero by the end of the pruning process. If the pruned model does not fit into the 320 KB memory of the ESP32, the values for initial and final sparsity are iteratively modified in 10\% increments until the final pruned model fits the microcontroller's memory. 

\section{Results}

 Average F1-scores (hypothesis 1) for the models are illustrated in Figure 6. The 1D CNN (0.64 $\mathrm{\pm}$ 0.17) and GRU (0.66 $\mathrm{\pm}$ 0.18) classifiers outperformed the RUSBoost (0.44 $\mathrm{\pm}$ 0.13), EasyEnsemble (0.44 $\mathrm{\pm}$ 0.13), and H2O-AutoML (0.22 $\mathrm{\pm}$ 0.07) models in terms of F1-scores (p $\mathrm{<}$ 0.01). Ten pairwise comparisons were made to include all combinations between each model (Table 4). Both 1D CNN and GRU outperformed RUSBoost, EasyEnsemble, and H2O-AutoML (p $\mathrm{<}$ 0.01), supporting our first hypothesis. Similarly, RUSBoost and EasyEnsemble outperformed the baseline H2O-AutoML (p $\mathrm{<}$ 0.01). The statistical testing showed that models belonging to the same category---neural networks (1D CNN and GRU, p = 0.42) and ensemble models (RUSBoost and EasyEnsemble, p = 0.48)---had statistically similar performance.

\begin{figure*}[ht]
    \centering
    \includegraphics[width=0.85\linewidth]{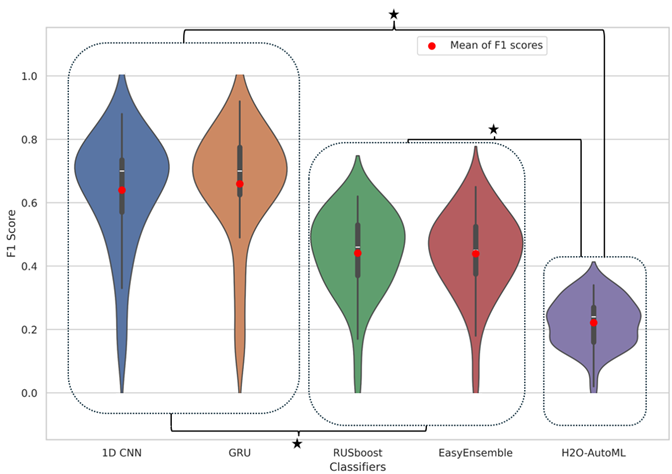}
    \caption{Comparison of F1-scores for five different models. Dotted lines distinguish models belonging to the same group: neural networks, ensembles, and the baseline H2O-AutoML. A star denotes a significant difference ($p < 0.01$) between the corresponding groups.
}
    \label{fig:6}
\end{figure*}

 The Wilcoxon signed-rank test indicated that the F1-scores of the 1D CNN and pruned 1D CNN (hypothesis 2) were statistically comparable (Table 4, p = 0.14; Fig. 7, left). However, the 1D CNN had lower memory sizes compared to the pruned 1D CNN (Table 4, p $\mathrm{<}$ 0.01; Fig. 7, right). 

\begin{figure*}[htbp]
    \centering
    \begin{minipage}[b]{0.45\textwidth}
        \centering
        \includegraphics[width=\textwidth]{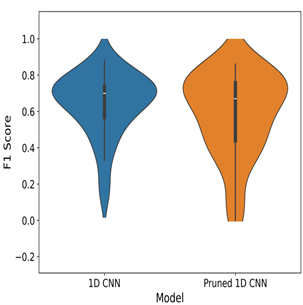}
    \end{minipage}
    \hfill
    \begin{minipage}[b]{0.45\textwidth}
        \centering
        \includegraphics[width=\textwidth]{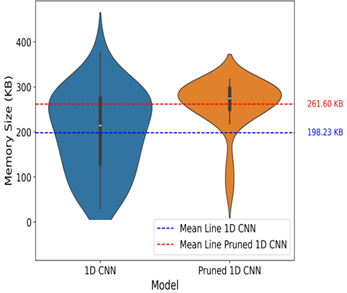}
    \end{minipage}
        \caption{Comparison of F1-scores (left) and memory sizes (right) for 1D CNN and pruned 1D CNN.}
        \label{fig:7}
\end{figure*}

 \begin{table*}[ht]
     \centering
 \caption{Details of ML model comparisons for three hypotheses using the Wilcoxon signed-rank test.}
\label{tab:4}
\begin{tabular}{@{}llll@{}}
\toprule
\textbf{Hypothesis}  & \textbf{Group 1} & \textbf{Group 2}       & \textbf{P-value}                       \\ \midrule
\multirow{11}{*}{H1} & 1D CNN           & GRU                    & \textbf{0.42}                          \\
                     & 1D CNN           & RUSBoost               & \textbf{$\boldsymbol{\mathrm{<}}$0.01} \\
                     & 1D CNN           & EasyEnsemble           & \textbf{$\boldsymbol{\mathrm{<}}$0.01} \\
                     & 1D CNN           & H2O-AutoML             & \textbf{$\boldsymbol{\mathrm{<}}$0.01} \\
                     & GRU              & RUSBoost               & \textbf{$\boldsymbol{\mathrm{<}}$0.01} \\
                     & GRU              & EasyEnsemble           & \textbf{$\boldsymbol{\mathrm{<}}$0.01} \\
                     & GRU              & H2O-AutoML             & \textbf{$\boldsymbol{\mathrm{<}}$0.01} \\
                     & RUSBoost         & EasyEnsemble           & \textbf{0.48}                          \\
                     & RUSBoost         & H2O-AutoML             & \textbf{$\boldsymbol{\mathrm{<}}$0.01} \\
                     & EasyEnsemble     & H2O-AutoML             & \textbf{$\boldsymbol{\mathrm{<}}$0.01} \\
                     & 1D CNN (memory)  & GRU (memory)           & \textbf{0.06}                          \\ \midrule
\multirow{2}{*}{H2}  & 1D CNN (memory)  & Pruned 1D CNN (memory) & \textbf{$\boldsymbol{\mathrm{<}}$0.01} \\
                     & 1D CNN           & Pruned 1D CNN          & \textbf{0.14}                          \\ \bottomrule
\end{tabular}

 \end{table*}

 A full report of the Macro F1-score, Precision, Macro Precision, Recall, and Macro Recall values is provided in Table A3 in the Appendix.  

\section{Discussion}

 The objectives of this study were to develop an AutoML framework and demonstrate its application in creating a fall detection algorithm tailored for the lower limb amputee community. This study introduced MicroNAS, an automated neural architecture search tool designed to optimize neural network models for deployment on resource-constrained microcontrollers, specifically the ESP32. The results of our hypothesis testing have provided promising results and highlighted several critical insights.

 \textbf{\textit{Hypothesis 1:}} Neural network models optimized with MicroNAS achieve higher performance, based on F1-score, compared to ensemble models or open source AutoML tools capable of handling class imbalance automatically and operating with minimal memory requirements for ESP32.  

 Hypothesis 1 compares the F1-score performance of multiple ML models and aims to determine the top-performing model for the ESP32. F1-scores demonstrated that the MicroNAS models, 1D CNN and GRU, achieved superior performance compared to RUSBoost, EasyEnsemble, and H2O-AutoML (Fig. 6).

 The higher F1-score performance of models created by MicroNAS (1D CNN and GRU) compared to other models can be attributed to three factors: 1) how the IMU data is introduced to the models, 2) how the data is processed, and 3) how class imbalance is handled. MicroNAS receives a window of 120 samples of IMU data, including a three-axis accelerometer and a three-axis gyroscope. It then extracts time-domain features from each window and transforms the representation of IMU data for the classification of falls and ADL. To handle class imbalance, MicroNAS incorporates greater weights for the fall class in the binary cross-entropy loss, ensuring that the misclassification of a fall is more severely penalized, thus affecting the model's parameter updates during the training phase. Consequently, by appropriately introducing the data, transforming its representation, and effectively addressing class imbalance through the weighting method, MicroNAS achieves higher F1-score performance.

 Ensemble models receive one sample of IMU data at a time and return the most common class across 120 consecutive predictions (a different method of windowing, Fig. 5). Unlike MicroNAS, ensemble models (RUSBoost and EasyEnsemble) do not extract features from IMU data. Manually extracting features by creating several mathematical functions may not be a feasible approach for deployment on a microcontroller with limited resources. Ensemble models address class imbalance by under sampling ADLs. However, the loss of samples due to under sampling during the training phase reduces the F1-score performance of ensemble models.

 Lastly, H2O AutoML receives one sample of IMU data at a time and makes a single prediction for each sample, without taking a majority vote from 120 predictions. To handle class imbalance, H2O AutoML over samples the fall class. However, over sampling can disrupt the temporal dependency between samples in IMU time series data and may introduce synthetic data points for the fall class that are not representative of real fall events, potentially misleading the ML model. Although we used H2O AutoML as prescribed, it was not appropriate for this context. This highlights one of the pitfalls of AutoML---misuse. The misuse of AutoML here demonstrates the importance of human oversight and domain expertise in ensuring that the models generated align with the true characteristics of the data (in our case, time series) and the techniques used for handling class imbalance are effective. The opaque nature of H2O AutoML did not allow us to optimize the model's hyperparameters based on the majority vote of 120 consecutive predictions during the training phase. H2O AutoML's poor F1-score performance underscores the importance of developing a customized AutoML pipeline for IMU data in fall detection, particularly when dealing with imbalanced data. As shown in Fig. 6, as the level of model customization decreases (MicroNAS models, Ensemble models, H2O AutoML), the F1-score performance also declines.

 The 1D CNN achieved a memory usage of 198.22 $\mathrm{\pm}$ 88.80 KB. In comparison, the GRU achieved a memory usage of 162.73 $\mathrm{\pm}$ 83.71 KB (Table A2 in the Appendix). The time complexity of 1D CNN and GRU in Big O notation is estimated by Equations 6 and 7, respectively.

 For 1D CNN,
\begin{equation}
    O\left(dFSCm\right)
\end{equation}
where $d$ is the number of convolutional layers, $F$ represents the number of filters in the $i^{th}$ layer, $S$  refers to the spatial dimensions of the filter, $C\ $indicates the number of input channels for the $i^{th}$ layer, and $m$ specifies the spatial dimensions of the output feature map [61].

 Time complexity of GRU is represented by 
\begin{equation}
    O\left(h\left(d+h\right)\right)
\end{equation}
where $h$ is the number of hidden cells and $d$ is the length of the input sequence [62]. An optimal architecture of the 1D CNN and GRU models created by MicroNAS for a randomly selected participant is provided in the Appendix for reference (Tables A4 and A5). 1D CNNs process different parts of the input windows in parallel, which may lead to faster inference speeds. This is particularly beneficial for the ESP32, which supports parallel processing with its dual-core CPU. In contrast, GRUs process data sequentially, potentially increasing both battery consumption and inference time. Ultimately, the end user should select the best model for deployment based on the specific needs of the FDS.

 \textbf{\textit{Hypothesis 2:}} Neural network model architectures generated by MicroNAS, without pruning, will achieve higher F1-scores and lower memory sizes compared to neural network models generated by NAS using magnitude-based weight pruning [18], to fit the memory constraint for ESP32. 

 Hypothesis 2 evaluates the effects of different search spaces and model development approaches on ML performance. The F1-score of the 1D CNN (0.64+/-0.17) was not statistically different from that of the pruned 1D CNN (0.58+/-0.23). Similar F1-score performance indicates that expanding the search space during the model development phase does not yield superior models compared to models created by MicroNAS.

 MicroNAS creates smaller 1D CNN models (198.22 $\mathrm{\pm}$ 89 KB) compared to the pruned 1D CNN created by NAS (261.60 $\mathrm{\pm}$ 56 KB) (p $\mathrm{<}$ 0.01, Table 4). The creation of smaller models conserves computational resources in two ways from an AutoML perspective:

\begin{enumerate}
\item  MicroNAS tunes 1D CNNs by developing smaller models during the training phase. It fully leverages the weight sharing attribute in 1D CNNs to reduce the number of parameters. MicroNAS utilizes a lower number of filters, reduces filter size, increases filter stride, and applies padding and pooling to create compact CNN models during the training phase.

\item  MicroNAS-developed models do not require a separate pruning stage. MicroNAS creates sparse 1D CNN models through train-time pruning, whereas the pruned 1D CNN applies pruning after the model has been fully trained (post-training pruning). MicroNAS employs train-time pruning by applying lasso regularization techniques (Table 2). Lasso regularization imposes a penalty on the loss function proportional to the absolute value of the weights, encouraging the model to retain only the most critical connections, thereby leading to a sparse representation. This approach contrasts with post-training pruning, which may not allow the model to adjust its learning based on the pruned architecture.
\end{enumerate}

 \subsection{Comparison to similar works}
 We acknowledge that comparing our ML models' F1-scores with those in previous studies is not appropriate due to differences in data sets. However, to illustrate the potential advantages offered by MicroNAS, we compare it to a similar FDS that developed deep learning models for pre-impact fall detection on the ARM 32-bit microcontroller [62] . The comparative study utilized a sampling frequency was 100 Hz, and used Keras Tuner for the model development. Sampling frequency of 50 Hz can also be used to save battery life [63]. The key differences between this previous study [62] and ours are as follows:

\begin{enumerate}
\item  Sensor placement: The previous study positioned the sensor on the waist. The waist is the steadiest part of the body, showing the least amount of noise and higher accuracy for FDS. In contrast, the thigh (upper shank) demonstrates a higher level of noise and consequently lower accuracy for FDS [63]. We specifically chose to place the IMU on the shank for amputees to integrate it with their prosthesis. Wearing sensors continuously on the body can be uncomfortable, particularly for amputees. To achieve effective real-time fall detection, it is crucial for the amputee to consistently wear the sensor [64], and the sensor on the prosthesis should cause no intrusion. 

\item  Class imbalance: The previous study had 60\% fall data (40\% ADL), whereas our study had only 1.7\% fall samples.

\item  Memory: The previous study had 512KB of memory for model creation, whereas we had 320KB.

\item  Performance: The previous study provided F1-score results for two participants in test data (1D CNN: 92.84, GRU: 71.92). Our average top F1-scores for two participants (participant 19 and participant A1) are 0.76 for the 1D CNN and 0.89 for the GRU (Table A2 in the Appendix). 
\end{enumerate}

\subsection{Significance}
 MicroNAS attained significantly higher F1-scores (0.66 $\mathrm{\pm}$ 0.18) compared to H2O AutoML (0.22 $\mathrm{\pm}$ 0.07). MicroNAS is an AutoML tool designed to process IMU data and create memory-efficient ML models (1D CNN and GRU) for the ESP32. Its open-source code enables other biomechanists to use and develop FDSs for their desired microcontroller. MicroNAS offers a user-friendly framework: biomechanists input the IMU data and the memory size of their microcontroller, initiate the optimization, and receive an exported model tailored to their microcontroller's memory requirements.

 MicroNAS provides detailed information about various aspects of model development to educate biomechanists on best practices and remove the opaque nature of ML libraries. These aspects include data splitting for human subject studies, data segmentation for IMU time series data, weighting methods and their effects on binary cross-entropy loss to handle class imbalance, search space design for varying architectures, hyperparameter range selection based on previous literature, the role of each hyperparameter, optimizers, and termination criteria based on available computational resources.

 The development of affordable and high-performance FDSs on microcontrollers by biomechanists can provide clinicians with objective insights into a patient's fall history. Clinicians can then use the fall history to develop intervention sessions and prescribe appropriate prosthetic devices.
\subsection{Limitations}
 Our FDS achieved its best performance with an F1-score of 0.92 with a GRU (Participant 19, Table 2 Appendix). The precision for the fall class was 0.87, and the recall was 0.97. To understand the real-world implications, consider an active individual with lower limb amputation using this FDS for 16 hours a day (excluding 8 hours of sleep and assuming two hours in motion for a moderately active individual). In this scenario, the system would generate 936 false alarms daily. Over a prolonged period, if 100 actual falls occur, the FDS would detect 97 of them.  To minimize false alarms to just one per day, the FDS would need to achieve an F1-score of approximately 0.99 for the fall class. Fall detection studies typically report metrics such as precision, recall, F1-score, and accuracy. However, it is crucial to focus on the F1-score specifically for the fall class rather than ADL. We achieved a perfect F1-score for the ADL class for participant 19 with GRU, which, although noteworthy, is less important in the context of fall detection.

 The current F1-scores are not sufficient for clinicians and commercial product implementation [62]. Our F1-scores were achieved with a single IMU sensor on the shank, data from only 35 participants, and a high imbalance ratio. The inclusion of another sensor (or placement at a different location) on the prosthesis could potentially increase the F1-score performance. In addition, we anticipate attaining higher F1-scores for FDS with the availability of a larger dataset for model training. The imbalance ratio was high in our dataset and rare events prediction like falls is inherently challenging. While our proposed weighting method could offset the effect of class imbalance, collecting more fall samples can further improve the FDS performance.

 Our FDS was developed based on laboratory data and the translation to the real world and functional use is still unknown. Age, previous fall experience, and velocity are contributing factors to real falls. The elderly are at a higher risk of falling compared to younger individuals. Moreover, the risk of a fall increases in individuals who have experienced a previous fall. Fall experience influences gait, and individuals at an increased risk of falling tend to walk more slowly [42]. Therefore, age and walking speed are influential in the actual risk of falling (which we did not control in this study), and accounting for these factors could likely enhance the performance of FDS in real-world scenarios. 

 Our laboratory-based fall simulation has fallen short of capturing the full scope of themes that emerged from the lower limb amputee community. A broader characterization of the physical environment (e.g., slippery surfaces, stairs, incline versus decline) and biomechanical factors (perturbations applied to the prosthetic leg during walking and transition activities, varying surface and terrain conditions, performing concurrent cognitive or physical tasks, and a variety of situations (e.g., fatigued, rushed)) is needed to provide a more comprehensive picture of which ground conditions contribute to falls or near-falls in the lower limb amputee population [65]. In addition, the definition of fall for a lower limb amputee should be considered: A fall is a loss of balance or sudden loss of support where the body lands on the ground, floor, or another object. A near-fall is a loss of balance where one catches oneself or recovers one's balance without landing on the ground, floor, or another object [66]. In our experimental fall simulation, participants completely landed on a fall pad during fall events. We did not simulate falls or near-falls that involved other objects. Additionally, the element of suddenness or unexpectedness was not fully present in our simulation. While participants wore glasses that blocked their peripheral vision when they hit the fall pad, they were mentally aware that a fall was about to occur.

 \subsection{Future work}
 In future research, we plan to assess the performance of our FDS, which was developed using laboratory-based data, against data collected from lower-limb amputees in real-world scenarios. Real-world falls may exhibit different data characteristics, and we aim to see if there is a significant statistical difference between the F1-scores of our model on test data in the lab with the test data in real-world scenarios. Currently, our ML models were developed with a majority of control participants. In the future, we will investigate whether an ML model developed with a majority of control participants will achieve similar performance (F1-score) for participants with lower limb amputation compared to control participants. In addition, we aim to develop MicroNAS current support and capabilities as follows:

\begin{enumerate}
\item  \textbf{Optimization Algorithms:} The MicroNAS pipeline optimizer currently uses random search. We plan to include evolutionary algorithms, such as genetic algorithms and particle swarm optimization, and assess their performance against the built-in optimizers in Keras Tuner, such as random search and Bayesian optimization.

\item  \textbf{Training Time Reduction:} For code execution, we utilized GPU processing; however, the development of our models was time-intensive. The GRU training process is sequential and slower compared to CNNs. To expedite training time for MicroNAS, in the future we plan on using adaptive subset selection [67]. Adaptive subset selection trains the MicroNAS models on a dynamically selected subset of the full training data, which is regularly updated during the training process to remain representative of the entire dataset. The objective is to preserve the model's performance while significantly reducing the computational resources required for training.

\item  \textbf{Handling Class Imbalance:} We used a weighting method for handling class imbalance. Other potential strategies for addressing class imbalance include the generation of synthetic data. While the Synthetic Minority Over-sampling Technique [68] is popular for oversampling, it does not account for the temporal dependency of data. Investigating synthetic data generation techniques that consider temporal dependencies in IMU data represents a promising direction for future research. 
\end{enumerate}

 \section{Conclusion}
 FDS development for individuals with lower limb amputations is crucial due to their increased risk of falling and the severe consequences of such falls. This research aimed to overcome the existing limitations in developing an FDS by addressing critical challenges such as the integration of ML techniques suitable for IMU sensors, class imbalance, and memory constraints of current microcontrollers. The contributions of this work are as follows:

\begin{enumerate}
\item  We collected simulated falls and ADL data from both control and lower limb amputee participants.

\item  We developed a neural architecture search tool (MicroNAS) for resource limited microcontrollers. MicroNAS uses the memory size of the microcontroller as a guide for model creation. MicroNAS effectively handles class imbalance and uses AutoML techniques to perform neural architecture search (1D CNN and GRU) for developing an FDS. 

\item  We demonstrated that MicroNAS saves computational resources by creating models for the ESP32 in a single stage without pruning.
\end{enumerate}

 MicroNAS's models surpass the performance of RUSBoost, EasyEnsemble, and H2O AutoML. MicroNAS neural network models achieved 68\% F1-score performance across all participants. The F1-score results were obtained using a single sensor located on the shank with 35 participants. A larger dataset will further refine our current models. Our models are deployable on a microcontroller with a mere 320KB of onboard memory. Biomechanists can utilize MicroNAS's open-source code to train neural network models for microcontrollers with diverse memory capacities.

 \section*{Acknowledgment}

 The authors acknowledge the support of Montana State University's Research Computing Infrastructure (RCI) for providing access to the Tempest high-performance computing cluster. Funding was provided by the CDMRP Grant W81XWH-20-1-0164.

 \section{Appendix}
\renewcommand{\thetable}{A\arabic{table}}
\setcounter{table}{0} 

\begin{table*}[ht]
    \centering
 \caption{Demographic information for participants.}
\label{tab:a1}
\begin{tabular}{|p{1.0in}|p{0.7in}|p{1.5in}|p{0.6in}|p{1in}|} \hline 
\textbf{Participant} & \textbf{Sex} & \textbf{Dominant or Amputation side} & \textbf{Age (years)} & \textbf{Amputation\newline Level} \\ \hline 
1 & F & R & 26 & - \\ \hline 
2 & F & R & 26 & - \\ \hline 
3 & F & R & 26 & - \\ \hline 
4 & F & R & 22 & - \\ \hline 
5 & F & R & 23 & - \\ \hline 
6 & F & R & 27 & - \\ \hline 
7 & M & R & 27 & - \\ \hline 
8 & F & R & 53 & - \\ \hline 
9 & M & L & 57 & - \\ \hline 
10 & F & R & 20 & - \\ \hline 
11 & M & R & 20 & - \\ \hline 
12 & M & R & 27 & - \\ \hline 
13 & F & L & 23 & - \\ \hline 
14 & F & R & 26 & - \\ \hline 
15 & F & R & 21 & - \\ \hline 
16 & M & R & 21 & - \\ \hline 
17 & F & R & 30 & - \\ \hline 
18 & M & R & 35 & - \\ \hline 
19 & F & R & 36 & - \\ \hline 
20 & M & L & 27 & - \\ \hline 
21 & F & R & 27 & - \\ \hline 
22 & F & R & 26 & - \\ \hline 
23 & F & R & 31 & - \\ \hline 
24 & M & R & 28 & - \\ \hline 
25 & F & L & 27 & - \\ \hline 
26 & F & R & 26 & - \\ \hline 
27 & F & R & 19 & - \\ \hline 
28 & F & R & 38 & - \\ \hline 
29 & M & R & 30 & - \\ \hline 
30 & F & R & 25 & - \\ \hline 
31 (A1) & M & L & 37 & Below Knee \\ \hline 
32 (A2) & F & L & 57 & Above Knee \\ \hline 
33 (A3) & M & R & 29 & Above Knee \\ \hline 
34 (A4) & M & Bilateral & 34 & Below Knee \\ \hline 
35 (A5) & M & R & 58 & Above Knee \\ \hline 
\end{tabular}

\end{table*}
 
\begin{table*}[ht]
    \centering
    
 \caption{F-score performance and memory size (KB) for all participants across different ML models}
\label{tab:a2}
\begin{tabular}{|p{0.7in}|p{0.3in}|p{0.4in}|p{0.3in}|p{0.46in}|p{0.5in}|p{0.4in}|p{0.4in}|p{0.3in}|p{0.4in}|p{0.4in}|} \hline 
Participant & 1D CNN & GRU & RUS\newline Boost & Easy \newline Ensemble &  H2O-AutoML &  Pruned 1D CNN &  1D CNN & GRU &  NAS 1D CNN & Pruned 1D CNN \\ \hline 
 & \multicolumn{6}{|p{2.1in}|}{F1 Score} & \multicolumn{4}{|p{1.3in}|}{Memory (KB)} \\ \hline 
1 & 0.72      & 0.65 & 0.43 & 0.45 & 0.24 & 0.80 & 84 & 184 & 1263 & 265 \\ \hline 
2 & 0.13 & 0.15 & 0.04 & 0.04 & 0.02 & 0.09 & 110 & 214 & 789 & 258 \\ \hline 
3 & 0.73 & 0.81 & 0.61 & 0.59 & 0.27 & 0.86 & 273 & 303 & 912 & 240 \\ \hline 
4 & 0.62 & 0.70 & 0.42 & 0.42 & 0.31 & 0.75 & 151 & 19 & 1243 & 266 \\ \hline 
5 & 0.73 & 0.79 & 0.35 & 0.37 & 0.25 & 0.80 & 125 & 82 & 1748 & 261 \\ \hline 
6 & 0.69 & 0.59 & 0.53 & 0.53 & 0.19 & 0.39 & 219 & 151 & 1925 & 289 \\ \hline 
7 & 0.83 & 0.58 & 0.37 & 0.37 & 0.12 & 0.68 & 109 & 149 & 1292 & 269 \\ \hline 
8 & 0.72 & 0.33 & 0.50 & 0.48 & 0.19 & 0.70 & 281 & 186 & 2156 & 316 \\ \hline 
9 & 0.57 & 0.16 & 0.41 & 0.44 & 0.15 & 0.42 & 309 & 86 & 1390 & 283 \\ \hline 
10 & 0.76 & 0.67 & 0.65 & 0.62 & 0.25 & 0.53 & 225 & 86 & 2144 & 315 \\ \hline 
11 & 0.84 & 0.73 & 0.35 & 0.35 & 0.27 & 0.64 & 275 & 219 & 757 & 295 \\ \hline 
12 & 0.72 & 0.36 & 0.63 & 0.60 & 0.30 & 0.85 & 261 & 274 & 981 & 266 \\ \hline 
13 & 0.74 & 0.84 & 0.38 & 0.37 & 0.14 & 0.82 & 142 & 230 & 302 & 136 \\ \hline 
14 & 0.57 & 0.79 & 0.40 & 0.40 & 0.25 & 0.81 & 176 & 160 & 170 & 65 \\ \hline 
15 & 0.84 & 0.75 & 0.48 & 0.46 & 0.34 & 0.75 & 291 & 157 & 572 & 302 \\ \hline 
16 & 0.72 & 0.73 & 0.40 & 0.59 & 0.30 & 0.64 & 206 & 350 & 688 & 296 \\ \hline 
17 & 0.83 & 0.78 & 0.44 & 0.57 & 0.17 & 0.81 & 311 & 134 & 1389 & 296 \\ \hline 
18 & 0.60 & 0.67 & 0.27 & 0.28 & 0.25 & 0.67 & 67 & 85 & 1095 & 292 \\ \hline 
19 & 0.88 & 0.92 & 0.52 & 0.54 & 0.24 & 0.80 & 235 & 120 & 1331 & 277 \\ \hline 
20 & 0.60 & 0.62 & 0.29 & 0.33 & 0.23 & 0.71 & 260 & 245 & 747 & 240 \\ \hline 
21 & 0.57 & 0.63 & 0.28 & 0.27 & 0.12 & 0.41 & 79 & 49 & 600 & 290 \\ \hline 
22 & 0.21 & 0.49 & 0.18 & 0.17 & 0.24 & 0.54 & 254 & 309 & 1367 & 288 \\ \hline 
23 & 0.33 & 0.83 & 0.53 & 0.51 & 0.32 & 0.29 & 58 & 211 & 1444 & 218 \\ \hline 
24 & 0.68 & 0.82 & 0.31 & 0.30 & 0.16 & 0.34 & 186 & 212 & 712 & 274 \\ \hline 
25 & 0.70 & 0.64 & 0.52 & 0.48 & 0.33 & 0.53 & 291 & 164 & 2763 & 295 \\ \hline 
26 & 0.48 & 0.71 & 0.40 & 0.39 & 0.24 & 0.67 & 29 & 184 & 2833 & 300 \\ \hline 
27 & 0.75 & 0.77 & 0.45 & 0.44 & 0.16 & 0.70 & 214 & 96 & 862 & 99 \\ \hline 
28 & 0.73 & 0.71 & 0.53 & 0.53 & 0.17 & 0.73 & 284 & 169 & 1502 & 249 \\ \hline 
29 & 0.73 & 0.68 & 0.51 & 0.50 & 0.15 & 0.76 & 305 & 278 & 1691 & 229 \\ \hline 
30 & 0.66 & 0.71 & 0.50 & 0.48 & 0.23 & 0.52 & 41 & 182 & 1111 & 303 \\ \hline 
31  & 0.64 & 0.86 & 0.61 & 0.60 & 0.33 & 0.00 & 168 & 106 & 1586 & 237 \\ \hline 
32  & 0.74 & 0.67 & 0.53 & 0.52 & 0.19 & 0.46 & 221 & 14 & 1798 & 251 \\ \hline 
33 & 0.41 & 0.59 & 0.59 & 0.60 & 0.16 & 0.50 & 132 & 24 & 1885 & 260 \\ \hline 
34  & 0.50 & 0.71 & 0.50 & 0.51 & 0.32 & 0.09 & 376 & 182 & 2137 & 314 \\ \hline 
35 & 0.42 & 0.64 & 0.46 & 0.33 & 0.15 & 0.36 & 173 & 66 & 531 & 304 \\ \hline 
\end{tabular}
\end{table*}

\begin{table*}[ht]
    \centering
     \caption{Comparison of models with different performance metrics on test data. Mean and standard deviation of each ML model represented for 35 participants.}
\label{tab:a3}
\begin{tabular}{p{1in}p{0.7in}p{0.7in}p{0.6in}p{1in}p{0.7in}p{1in}} \hline 
\textbf{Model} & \textbf{F1 } & \textbf{Macro F1} & \textbf{Precison} & \textbf{Macro Precison} & \textbf{Recall} & \textbf{Macro Recall} \\ \hline 
\textbf{1D CNN} & ~0.64+/-0.17 & ~0.82+/-0.09 & ~0.67+/-0.20 & ~0.83+/-0.10 & ~0.70+/-0.20 & ~0.85+/-0.10 \\ \hline 
\textbf{GRU} & ~0.66+/-0.18 & ~0.82+/-0.09 & 0.65+/-0.19~ & ~0.82+/-0.09 & ~0.76+/-0.21 & ~0.87+/-0.11 \\ \hline 
\textbf{RUSBoost} & ~0.44+/-0.13 & ~0.70+/-0.11 & ~0.30+/-0.10 & ~0.65+/-0.05 & ~0.94+/-0.06 & ~0.94+/-0.07 \\ \hline 
\textbf{EasyEnsemble} & ~0.44+/-0.13 & ~0.70+/-0.11 & ~0.29+/-0.10 & ~0.65+/-0.05 & ~0.95+/-0.05 & ~0.94+/-0.07 \\ \hline 
\textbf{H2O AutoML} & ~0.22+/-0.07 & ~0.59+/-0.07 & ~0.28+/-0.14 & ~0.63+/-0.07 & ~0.25+/-0.13 & ~0.60+/-0.07 \\ \hline 
\textbf{Pruned 1D CNN} & ~0.58+/-0.23 & ~0.82+/-0.09 & ~0.66+/-0.18 & ~0.82+/-0.09 & ~0.75+/-0.20 & ~0.87+/-0.10 \\ \hline 
\end{tabular}

\end{table*}

 \begin{table*}[ht]
     \centering
\caption{1D CNN Optimal Architecture Diagram Created by MicroNAS}
\label{tab:a4}
\begin{tabular}{@{}lll@{}}
\toprule
Layer (type)             & Output Shape  & Param \#          \\ \midrule
Batch Normalization      & None, 120, 6  & 24                \\
Conv1D                   & None, 60, 141 & 6909              \\
Batch Normalization      & None, 60, 141 & 564               \\
Pooling Layer1D          & None, 30, 141 & 0                 \\
Conv1D                   & None, 15, 31  & 4402              \\
Pooling Layer1D          & None, 7, 31   & 0                 \\
Conv1D                   & None, 7, 291  & 27354             \\
Batch Normalization      & None, 7, 291  & 1164              \\
Pooling Layer1D          & None, 3, 291  & 0                 \\
Global Average Pooling1D & None, 291     & 0                 \\
Flatten                  & None, 291     & 0                 \\
Batch Normalization      & None, 291     & 1164              \\
Dense                    & None, 1       & 292               \\ \midrule
Total Parameters         &               & 41873(163.57 KB)  \\
Trainable Parameters     &               & 40415 (157.87 KB) \\
Non-trainable Parameters &               & 1458 (5.70 KB)    \\ \bottomrule
\end{tabular}

 \end{table*}
 
\begin{table*}[ht]
    \centering
    \caption{GRU Optimal Architecture Diagram Created by MicroNAS}
\label{tab:a5}
\begin{tabular}{@{}lll@{}}
\toprule
Layer (type)             & Output Shape  & Param \#           \\ \midrule
Batch Normalization      & None, 120, 6  & 24                 \\
GRU                      & None, 120, 30 & 3420               \\
Batch Normalization      & None, 120, 30 & 120                \\
Global Average Pooling1D & None, 30      & 0                  \\
Flatten                  & None, 30      & 0                  \\
Batch Normalization      & None, 30      & 120                \\
Dropout                  & None, 30      & 0                  \\
Batch Normalization      & None, 30      & 120                \\
Dropout                  & None, 30      & 0                  \\
Batch Normalization      & None, 30      & 120                \\
Dense                    & None, 1003    & 31093              \\
Dropout                  & None, 1003    & 0                  \\
Dense                    & None, 1       & 1004               \\
Total Parameters         &               & 36021(140.71 KB)   \\ \midrule
Trainable Parameters     &               & 35769 (139.72 KB)  \\
Non-trainable Parameters &               & 252 (1008.00 Byte) \\ \bottomrule
\end{tabular}

\end{table*}

\end{document}